\documentclass[11pt]{article}

\usepackage[preprint]{acl}

\usepackage{times}
\usepackage{latexsym}

\usepackage[T1]{fontenc}

\usepackage[utf8]{inputenc}

\usepackage{microtype}

\usepackage{inconsolata}

\usepackage{graphicx}

\usepackage{amsmath}
\usepackage{amsfonts}
\usepackage{booktabs}
\usepackage{multirow}
\usepackage{makecell}
\usepackage[most]{tcolorbox}
\usepackage[ruled,linesnumbered,noend]{algorithm2e}
\usepackage{algpseudocode}
\usepackage{nccmath}
\usepackage[most]{tcolorbox}
\usepackage{csquotes}
\usepackage{anyfontsize}
\usepackage{comment}
\usepackage{float}
\usepackage{cuted}
\usepackage{tikz}
\usepackage{listings,multicol}
\usepackage[
    framemethod=tikz,
    skipbelow=\topskip,
    skipabove=\topskip
]{mdframed}
\newtcolorbox[list inside=prompt,auto counter,number within=section]{prompt}[1][]{
    colbacktitle=black,
    colback=white,
    coltitle=white,
    fontupper=\footnotesize,
    boxsep=5pt,
    left=0pt,
    right=0pt,
    top=0pt,
    bottom=0pt,
    boxrule=1pt,
    #1,
}
\usepackage{ulem}
\usepackage{paralist}
\usepackage{pifont}
\usepackage{marvosym}

\title{HeteroRAG: A Heterogeneous Retrieval-Augmented Generation Framework for Medical Vision Language Tasks}

\author{
    Zhe~Chen$^{1,3}$,
    Yusheng~Liao$^{1,3}$,
    Zhiyuan~Zhu$^{1}$,
    Haolin~Li$^{2,3}$,
    Hongcheng~Liu$^{1}$
    \\
    \bf
    Yanfeng~Wang$^{1,3}$,
    Yu~Wang$^{1,3}$\textsuperscript{\Letter} \vspace{0.6mm}
    \\ 
    \begin{tabular}{c} 
    $^1$Shanghai Jiao Tong University ~~~
    $^2$Fudan University \\
    $^3$Shanghai Artificial Intelligence Laboratory \vspace{0.6mm}\\
    \end{tabular}
    \\ 
    \begin{tabular}{c}
    \texttt{\{chenzhe2018,yuwangsjtu\}@sjtu.edu.cn}
    \end{tabular}
}

\begin{document}
\maketitle

\renewcommand{\thefootnote}{}
\footnotetext{\Letter: Corresponding author.}
\renewcommand{\thefootnote}{\arabic{footnote}} 

\begin{abstract}

Medical large vision-language Models (Med-LVLMs) have shown promise in clinical applications but suffer from factual inaccuracies and unreliable outputs, posing risks in real-world diagnostics. While RAG has emerged as a potential solution, current medical multimodal RAG systems are unable to perform effective retrieval across heterogeneous sources. The irrelevance of retrieved reports undermines the factuality of analysis, while insufficient knowledge affects the credibility of clinical decision-making. To bridge the research gap, we construct MedAtlas, which includes extensive multimodal report repositories and diverse text corpora. Based on it, we present HeteroRAG, a novel framework that enhances Med-LVLMs through heterogeneous knowledge sources. The framework introduces Modality-specific CLIPs for effective report retrieval and a Multi-corpora Query Generator for tailoring queries to diverse corpora. Incorporating knowledge from such multifaceted sources, Heterogeneous Knowledge Preference Tuning is performed to achieve cross-modality and multi-source knowledge alignment. Extensive experiments across 11 datasets and 3 modalities demonstrate that HeteroRAG achieves state-of-the-art performance in most medical vision language benchmarks, significantly improving factual accuracy and reliability of Med-LVLMs\footnote{Project website: \url{https://github.com/Jack-ZC8/HeteroRAG-Med}}. 
\end{abstract}

\section{Introduction}

Large vision-language models (LVLMs) have made significant strides in integrating multimodal information and generating natural responses \cite{Chen2024a, Chen2024c, Liu2024a, Comanici2025, Bai2025}. Similarly, medical LVLMs (Med-LVLMs) show increasing promise for multimodal diagnosis and clinical decision support \cite{Chen2024, Liu2024b, Lin2025, Xu2025}. However, despite these advances, current Med-LVLMs still struggle with critical challenges related to factual accuracy and reliability \cite{Sun2025, Xia2025}, which is essential in diagnosis \cite{Xiong2024, Li2025a}. This limitation poses serious risks in medical applications, where errors could lead to misdiagnosis or harmful treatment recommendations.

\begin{figure}[t]
    \centering
    \includegraphics[width=\linewidth]{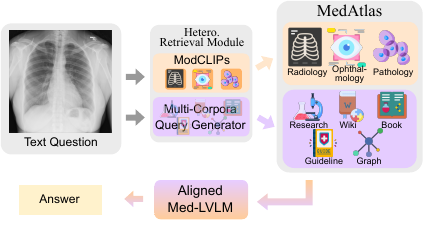}
    \caption{Overview of HeteroRAG Framework. The HRM retrieves reports and documents from MedAtlas for a knowledge-aligned Med-LVLM.}
    \label{fig:overview}
\end{figure}

To mitigate these limitations, recent scholarly efforts have prioritized multimodal retrieval-augmented generation (MMRAG) frameworks, which augment Med-LVLMs with medical knowledge to enhance diagnostic accuracy and epistemic reliability \cite{Ranjit2023, Sun2025, Choi2025, Shaaban2025}. 
Predominant methodologies employ multimodal retrievers, such as the medical modality-aware CLIP models, to retrieve relevant reports using input images, owing to the strong semantic similarity between medical imaging and textual reports \cite{Sun2025, Xia2024, Xia2025}. \textbf{However, the training data used to enhance the modality awareness of these retrievers is typically limited to the training splits of only a few datasets.} This constraint leads to inferior retrieval performance and irrelevant retrieved reports, undermining the factuality of the Med-LVLMs.

Moreover, medical corpora, such as research articles and clinical guidelines, are crucial for enhancing the reliability of Med-LVLMs. However, the multimodal retrievers mentioned above fail when applied to corpora, since these corpora lack direct visual semantics and exhibit diverse linguistic characteristics.
Current efforts \cite{Wu2025, Hamza2025} typically conduct cross-modality document retrieval using the original multimodal query. \textbf{Though straightforward, they fail in corpus-specific retrieval as they neglect the alignment between queries and corpus linguistic characteristics as noted above.} MIRA~\cite{Wang2025a}, which employs zero-shot LLM-rewritten queries, still lacks this customizability due to limited information presented in the rewriting prompt. \textbf{In summary, current medical MMRAG works remain ineffective in retrieving across heterogeneous imaging reports and literature corpora, resulting in a significant knowledge gap.}

A key bottleneck in addressing these limitations is the lack of a diverse and sufficient knowledge base. To fill the gap, we construct MedAtlas, which comprises broad multimodal report repositories and rich text corpora. The report repositories contain image-text reports in radiology, ophthalmology, and pathology. The text corpora are compiled from research articles, Wikipedia entries, medical textbooks, clinical guidelines, and knowledge graphs.

Building upon MedAtlas, we propose HeteroRAG, a framework designed to significantly enhance the factual accuracy and reliability of Med-LVLMs. As illustrated in Figure~\ref{fig:overview}, we develop the Heterogeneous Retrieval Module (HRM), which integrates Modality-specific CLIPs (ModCLIPs) and a Multi-corpora Query Generator (MQG). ModCLIPs are trained on large-scale data to ensure effective cross-modality report retrieval. The MQG module is trained in two stages to capture corpus-specific characteristics and generate tailored queries.
Finally, we propose the Heterogeneous Knowledge Preference Tuning (HKPT) method to achieve two types of alignment: (1) cross-modality alignment, which aligns visual inputs with retrieved textual content; and (2) multi-source knowledge alignment, which aligns the model’s internal knowledge with external knowledge from diverse sources.

We evaluate HeteroRAG on medical visual question answering and report generation tasks across 3 modalities and 11 datasets. Empirical results show that our framework achieves state-of-the-art performance on most benchmarks, demonstrating its strong factuality and reliability. Notably, HeteroRAG surpasses public Med-LVLMs, which contain 4–5× parameters, highlighting the value of effective knowledge integration and alignment.

Our contributions are summarized as follows:
\begin{itemize}
\item 
We introduce MedAtlas, a newly curated, comprehensive medical database that provides rich multimodal knowledge for Med-LVLMs and establishes a robust foundation for medical MMRAG research.

\item 
Leveraging MedAtlas, we propose HeteroRAG, a novel medical MMRAG framework that performs accurate heterogeneous knowledge retrieval and fine-grained knowledge alignment.

\item 
Extensive experiments validate HeteroRAG's capability to precisely retrieve and effectively integrate multi-source knowledge, demonstrating SOTA performance across most benchmarks. The framework also consistently outperforms substantially larger Med-LVLMs and establishes a trustworthy and reliable foundation for medical applications.
\end{itemize}

\section{Related Work}

\subsection{Report Retrieval in Medical MMRAG}

Existing medical MMRAG approaches primarily utilize the medical images to retrieve relevant reports \cite{He2024, Sun2025, Xia2024, Xia2025}. For instance, FactMM-RAG~\cite{Sun2025} enhances report generation by incorporating high-quality reference reports. Similarly, RULE~\cite{Xia2024} and MMed-RAG~\cite{Xia2025} integrate reference reports and employ preference fine-tuning to improve model utilization of retrieved reports. Although these approaches improve the factual accuracy of responses, they neglect the retrieval of medical documents, which are crucial for Med-LVLM's reliable inference.

\subsection{Document Retrieval in Medical MMRAG} 

Acknowledging the limitations of report-only retrieval, recent medical MMRAG studies have increasingly emphasized medical documents as knowledge sources \cite{Choi2025, Shaaban2025, Wu2025, Hamza2025}. Among them, MKGF~\cite{Wu2025} and K-LLaVA~\cite{Hamza2025} both employ multimodal retrievers to fetch documents from the database, aiming to mitigate hallucination issues in language models. ChatCAD+~\cite{Zhao2024a} and MIRA~\cite{Wang2025a} utilize a zero-shot query rewriting module for retrieval. Nevertheless, these retrieval methods overlook the substantial content differences among various corpora, lacking corpus-specific retrieval mechanisms.  While corpus-specific query generation has shown promise in text-only RAG~\cite{Chen2025}, such methods lack cross-modal perception of visual evidence, and applying them to multimodal clinical scenarios remains an open challenge.

\section{MedAtlas Knowledge Base}

\begin{figure*}[t]
    \centering
    \includegraphics[width=1\linewidth]{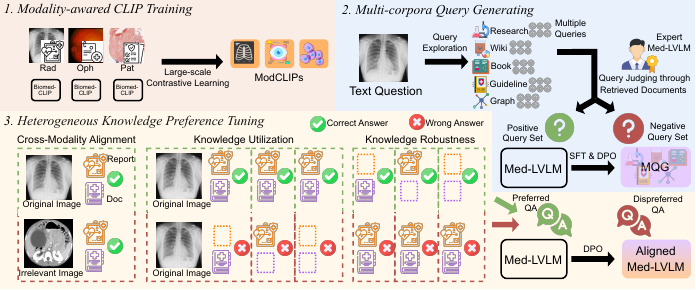}
    \caption{Overview of HeteroRAG framework. It introduces the Modality-specific CLIPs for effective report retrieval. Then, the Multi-Corpora Query Generator is developed for tailored retrieval for different corpora. Finally, HKPT is conducted to achieve the cross-modality and multi-source knowledge alignment.}
    \label{fig:method}
\end{figure*}

The MedAtlas knowledge base comprises comprehensive multimodal report repositories covering three modalities and rich textual corpora from five distinct sources.

\subsection{Multimodal Report Repository}

Existing medical image-report repositories are limited in scale and diversity, typically being derived from the training splits of a few report generation datasets \cite{Sun2025, Xia2024, Xia2025}. To address this issue, we collect image-report pairs from a wide range of datasets. Specifically, the Radiology subset includes 1,104,313 pairs from 6 datasets; the Ophthalmology subset includes 111,991 pairs from 5 datasets; and the Pathology subset includes 1,514,058 pairs from 5 datasets.
To ensure data quality, duplicate pairs are removed using the image perceptual hashing algorithm~\cite{Du2020}. More details are provided in Appendix~\ref{app:medatlas}. For the retrieval method, we use images as queries and reports in the library as keys to retrieve the top-k reports, following \citet{Sun2025, Xia2024, Xia2025}.

\subsection{Textual Corpora}

To ensure the richness, we collect corpora from representative sources, following \citet{Xiong2024, Chen2025}. The \textbf{Research} corpus is drawn from the 2025 PubMed Annual Baseline. The \textbf{Wiki} corpus is collected from the Wikipedia dumps. The \textbf{Book} corpus contains E-books, MedQA Textbook, and StatPearls, providing foundational medical knowledge \cite{Xiong2024, Fan2025, Chen2025}. The \textbf{Guideline} corpus contains clinical guidelines crawled from authoritative websites following \citet{Chen2023}. For the above four corpora, they are chunked into chunks of no more than 1000 characters, with an overlap of 200 characters following \citet{Xiong2024}.

For the retrieval of unstructured corpora, each query is formatted as ``query'', and the MedCPT models \cite{Jin2023} are used for vector search and reranking. For the structured Graph corpus, each query is formatted as ``query\_term, query\_relation''. Given the ``query\_term'', its definition and all one-hop relationships are retrieved, followed by filtering relevant relationships by reranking with ``query\_relation'' \cite{Yang2024}.

\section{HeteroRAG Framework}

In this section, we present the HeteroRAG framework, as illustrated in Figure~\ref{fig:method}. First, we introduce Modality-specific CLIPs (ModCLIPs), which are trained on large-scale image-text pairs for accurate report retrieval. Next, a Multi-Corpora Query Generator (MQG) is developed to enable tailored retrieval for multimodal questions based on corpus characteristics. Finally, we propose a Heterogeneous Knowledge Preference Tuning (HKPT) method to realize cross-modality and multi-source knowledge alignment.

\subsection{Modality-specific CLIPs}\label{sec:method_clips}

The ModCLIPs are initialized from BiomedCLIP \cite{Zhang2023}. For each modality, the report retrieval base is independently split into training, development, and test sets to fine-tune CLIP models, following \citet{Xia2024, Xia2025}. Specifically, all samples of each modality are randomly split into 2000 development samples, 2000 test samples, and the remainder for training. This results in 1.10M image-text training pairs in radiology, 0.11M in ophthalmology, and 1.51M in pathology. Contrastive learning \cite{Radford2021} is performed on single-modality image-text pairs for each ModCLIP. Compared to previous work \cite{Sun2025, Xia2024, Xia2025}, which relied solely on training splits from a limited number of datasets, the significantly scaled-up training data enables more accurate cross-modal report retrieval.

\subsection{Multi-corpora Query Generator}

For each multimodal question including the image $v$ and text question $t$, the module generates query set for each corpus $Q = \{ (i, j, q_{j}^{i}) \mid i = 1, 2, ..., N_{C}, \; j=1,2,..., N^{i}_{q}\}$, where $q_{j}^{i}$ denotes the $j_{th}$ query for the $i_{th}$ corpus, $N_{C}$ denotes the number of corpora, and $N^{i}_{q}$ denotes the number of queries for the $i_{th}$ corpus. Each query is then used to retrieve documents that collectively support answering $(v,t)$. Inspired by and extending prior work on text-only tailored query generation \cite{Chen2025}, we design the following pipeline for the medical MMRAG tasks.

Since annotations for documents supporting medical multimodal questions are generally unavailable, we use the LVLMs to generate proxy labels as inspired by \citet{Gu2024, Li2025, Chen2025}. We begin with a query exploration phase to identify potential retrieval strategies. Lingshu-32B~\cite{Xu2025} is selected as it consistently achieves strong performances across medical vision language tasks, and it is quantized with AWQ for inference efficiency. We prompt the expert Med-LVLM to generate multiple queries for each source, with the prompt shown in Prompt~\ref{prompt:query_exploration}. The prompts are designed to encourage intra-corpus diversity and align with the characteristics of corpora. To control the cost, the number of exploration queries per corpus is fixed to 6. Subsequently, the same expert model evaluates the documents retrieved by each query by judging whether they support the reference answer, with the prompt shown in Prompt~\ref{prompt:query_judging}. To evaluate the annotation quality, manual judging is conducted on a 500-item subset\footnote{A sample size of 500 was determined using Cochran's Formula, yielding a margin of error of ±4.38\% at a 95\% confidence level.} by medical researchers, also following the instructions in Prompt~\ref{prompt:query_judging}. They were informed that the results would be used only for research, and their participation was voluntary. The results show that Lingshu-32B achieves an accuracy of 0.836 and an F1 score of 0.855 against manual expert judgments, demonstrating the reliability of VLM-as-a-judge. Based on Lingshu-32B's judgments, queries are categorized as either positive, denoted $q_{w}$, or negative, denoted $q_{l}$.

For each corpus, we select up to $N^{i}_{q}$ instances of $q_{w}$ and $q_{l}$ to form positive queries $Q_{w}$ and negative queries $Q_{l}$, respectively. A two-stage training strategy is applied to MQG. First, supervised fine-tuning (SFT) is performed:

\begin{equation}
    \mathcal{L}_{\text{SFT}} = - \mathbb{E}_{\left(v,t, Q_w\right) \sim \mathcal{D}_w} \log \mathcal{M}_{\theta} \left(Q_w \mid v,t\right).
\end{equation}

Then, direct preference optimization (DPO) is applied to further align the retrieval strategies with corpora:
\begin{equation}
\begin{array}{l}
\mathcal{L}_{\text{DPO}}(\mathcal{M}_\theta; \mathcal{M}_{\text{ref}}) = -\mathbb{E}_{(v,t,Q_w,Q_l) \sim \mathcal{D}_{wl}} \\
\left[ \log \sigma
\left(
\beta \log \frac{\mathcal{M}_\theta(Q_w | v,t)}{\mathcal{M}_{\text{ref}}(Q_w | v,t)}
- \beta \log \frac{\mathcal{M}_\theta(Q_l | v,t)}{\mathcal{M}_{\text{ref}}(Q_l | v,t)}
\right) 
\right].
\end{array}
\end{equation}

\begin{algorithm}[t]
\small
\caption{Heterogeneous Knowledge Preference Tuning (HKPT)}
\LinesNumbered
\label{ag:hkpt}
\KwIn{$\mathcal{D}=\{v^{i},t^{i},K^{i},y^{i}\}_{i=1}^N$: Training dataset; $K=\{k_r,k_d\}$: Retrieved knowledge; $\mathcal{M}_\theta$: Med-LVLM; $\mathcal{D}_{cm}, \mathcal{D}_{mk}$: Preference datasets.}
\KwOut{$\mathcal{M}$: Preference tuned model.}
Initialize $\mathcal{D}_{cm}, \mathcal{D}_{mk}$ with empty sets\\
\ForEach{$(v,t,K,y) \in \mathcal{D}$}{
    Retrieve the image $v^*$ irrelevant to $v$ \\
    $\triangleright$ \texttt{Cross-Modality Alignment} \\
    \If{$\mathcal{M}(v,t,K)=y$ and $\mathcal{M}(v^*,t)\neq y$ and $\mathcal{M}(v^*,t,K)=y$}{
        $x_{w} \leftarrow (v,t,K)$; \quad $y_{w} \leftarrow y$ \\
        $x_{l} \leftarrow (v^*,t,K)$; \quad $y_{l} \leftarrow \mathcal{M}(v^*,t,K)$\\
        Put $\{x_{w},x_{l},y_{w},y_{l}\}$ into $\mathcal{D}_{cm}$ \\
      }
    $\triangleright$ \texttt{Multi-Source Knowledge Alignment} \\
    \ForEach{$k \in \{\{k_r\},\{k_d\},\{k_r,k_d\}\} $}{
        $\triangleright$ \texttt{Knowledge Utilization} \\
        \If{$\mathcal{M}(v,t,K)=y$ and $\mathcal{M}(v,t,K_{\setminus k})\neq y$}{
            $x_{w} \leftarrow (v,t,K)$; \quad $y_{w} \leftarrow y$ \\
            $x_{l} \leftarrow (v,t,K)$; \quad $y_{l} \leftarrow \mathcal{M}(v,t,K_{\setminus k}) $ \\
            Put $\{x_{w},x_{l},y_{w},y_{l}\}$ into $\mathcal{D}_{mk}$ \\
          }
        $\triangleright$ \texttt{Knowledge Robustness} \\
        \If{$\mathcal{M}(v,t,K_{\setminus k})=y$ and $\mathcal{M}(v,t,K)\neq y$}{
            $x_{w} \leftarrow (v,t,K)$; \quad $y_{w} \leftarrow y$ \\
            $x_{l} \leftarrow (v,t,K)$; \quad $y_{l} \leftarrow \mathcal{M}(v,t,K)$ \\
            Put $\{x_{w},x_{l},y_{w},y_{l}\}$ into $\mathcal{D}_{mk}$ \\
          }
      }
    }
\ForEach{$(x_{w},x_{l},y_{w},y_{l}) \in \mathcal{D}_{cm} \cup \mathcal{D}_{mk}$}{
    Compute the loss and update $\mathcal{M}$ following Eq.~\ref{eq:hkpt}
}
\end{algorithm}

\subsection{Heterogeneous Knowledge Preference Tuning} \label{sec:hkpt}

Despite retrieving relevant reports and reliable documents, Med-LVLMs still suffer from severe knowledge misalignment issues. Inspired by RULE~\cite{Xia2024} and MMed-RAG~\cite{Xia2025}, which introduce the preference fine-tuning strategy for aligning Med-LVLMs with external reports, we propose Heterogeneous Knowledge Preference Tuning (HKPT) to enable alignment with knowledge from more sources. The HKPT process is detailed in Algorithm~\ref{ag:hkpt}.

\paragraph{Cross-Modality Alignment.}
The incorporation of external knowledge may cause Med-LVLM to ignore visual information and directly copy retrieved contents \cite{Xia2025}. To mitigate this, we construct preference pairs from the training set to improve modality alignment. Each training sample is denoted as $\{v, t, K, y\}$, where $v$ is the medical image, $t$ is the text question, $K$ is the retrieved knowledge (including reports $k_{r}$ and documents $k_{d}$), and $y$ is the gold answer. For each $v$, we retrieve the least similar image based on the ModCLIP image feature from the same modality training samples as an irrelevant image~$v^*$. Preferred responses are selected when $\mathcal{M}$ correctly answers using $v$, while dispreferred ones are selected when $\mathcal{M}$ correctly answers using irrelevant $v^*$, indicating that $\mathcal{M}$ ignores $v$ and relies solely on $K$. For open-ended generation tasks, correctness is defined as the average metric exceeding a threshold $\alpha_{r}$. The criterion also applies below. This process forms the preference dataset $\mathcal{D}_{cm}$.

\begin{table*}[htbp]
    \centering
    \setlength{\tabcolsep}{1mm}
    \resizebox{\linewidth}{!}{
    \begin{tabular}{l c ccc cc ccc}
    \toprule[1pt]
        \multirow{2}{*}{\textbf{Methods}} & \multirow{2}{*}{\textbf{Retrieval}} &\multicolumn{3}{c}{\textbf{Radiology}} & \multicolumn{2}{c}{\textbf{Ophthalmology}} & \multicolumn{3}{c}{\textbf{Pathology}} \\ 
        \cmidrule(r){3-5} \cmidrule(r){6-7} \cmidrule(r){8-10}     
        & & VQA-RAD & SLAKE & OMVQA-Rad$^\dagger$ & DME-VQA & OMVQA-Oph$^\dagger$ & PathMMU & PathVQA & Quilt-VQA$^\dagger$ \\
        \midrule
Original & - & 72.79 & 83.65 & 74.92 & 81.92 & 80.83 & 57.36 & 77.38 & 49.27 \\
\midrule
Beam Search & - & 73.16 & 83.17 & 75.17 & 81.92 & 80.75 & 57.69 & 76.76 & 45.77 \\
DoLa & - & 76.47 & 82.21 & 73.67 & 79.33 & 80.17 & 55.35 & 80.86 & 66.76 \\
VCD & - & 71.69 & 81.73 & 73.42 & 80.32 & 80.42 & 56.86 & 75.05 & 52.77 \\
AVISC & - & 73.16 & 83.65 & 75.50 & 81.62 & 81.33 & 57.69 & 77.41 & 50.15 \\
M3ID & - & 72.79 & 82.93 & 74.83 & 81.85 & 80.83 & 57.53 & 76.59 & 45.77 \\
\midrule
MedDr & Report & 73.16 & 83.65 & 75.08 & 79.71 & 79.50 & 61.20 & 77.74 & 53.35 \\
FactMM-RAG & Report & 76.84 & 83.89 & 75.58 & 81.92 & 81.50 & 73.58 & \underline{91.98} & \underline{69.68} \\
RULE & Report & 73.16 & 84.38 & 74.67 & 82.61 & 79.50 & 65.72 & 81.95 & 60.35 \\
MMed-RAG & Report & 75.74 & \underline{86.06} & 76.33 & 80.70 & 79.08 & 68.06 & 85.67 & 67.35 \\
\midrule
MKGF & Doc & 74.63 & 84.86 & 74.25 & 82.07 & 82.33 & 66.22 & 80.06 & 58.02 \\
K-LLaVA & Doc & \underline{77.21} & 84.62 & 76.00 & \underline{88.48} & \underline{83.75} & 73.75 & 87.76 & 61.81 \\
\midrule
MIRA & Report+Doc & 76.84 & 84.38 & \underline{76.58} & 87.95 & 82.50 & \underline{74.25} & \textbf{92.10} & 68.80 \\
\textbf{HeteroRAG (Ours)} & Report+Doc & \textbf{81.99} & \textbf{87.50} & \textbf{80.42} & \textbf{88.56} & \textbf{86.00} & \textbf{75.59} & 90.83 & \textbf{72.89} \\

    \bottomrule[1pt]
    \end{tabular}
    }
    \caption{Model performance of different methods based on Lingshu-7B on the medical VQA task. The best results and second-best results are highlighted in \textbf{bold} and \underline{underlined}, respectively. $\dagger$: out-of-distribution datasets.}
    \label{tab:main_vqa}
\end{table*}

\paragraph{Multi-Source Knowledge Alignment.}
To improve $\mathcal{M}$'s alignment with external knowledge $K$, which includes reports $k_{r}$ and documents $k_{d}$, we design preference pairs from two aspects: \textbf{knowledge utilization and robustness}. Taking $k_{r}$ as an example: For knowledge utilization, preferred responses are selected when $\mathcal{M}$ correctly answers by properly using $k_{r}$, while dispreferred ones are selected when $\mathcal{M}$ fails without $k_{r}$. For knowledge robustness, preferred responses are selected when $\mathcal{M}$ correctly answers without $k_{r}$, while dispreferred ones are selected when $\mathcal{M}$ misuses $k_r$ and produces incorrect answers. The dual-aspect strategy is also applied to $k_d$, and a combination of $k_r$ and $k_d$, ensuring fine-grained alignment across all knowledge sources.

The resulting $D_{mk}$, together with $D_{cm}$, are employed in HKPT, enabling unified alignment across modalities and knowledge sources:
\begin{equation}\label{eq:hkpt}
\begin{array}{l}
\mathcal{L}_{\text{HKPT}}(\mathcal{M}_{\theta'}; \mathcal{M}_{\text{ref}'}) = -\mathbb{E}_{(x_w,x_l,y_w,y_l) \sim \mathcal{D}_{cm} \cup \mathcal{D}_{mk}} \\
\left[ \log \sigma
\left(
\beta \log \frac{\mathcal{M}_{\theta'}(y_w | x_w)}{\mathcal{M}_{\text{ref}'}(y_w | x_w)}
- \beta \log \frac{\mathcal{M}_{\theta'}(y_l | x_l)}{\mathcal{M}_{\text{ref}'}(y_l | x_l)}
\right) 
\right].
\end{array}
\end{equation}

\section{Experiments}

\subsection{Experimental Setups}

\paragraph{Datasets and Metrics.}

The medical VQA datasets include VQA-RAD~\cite{Lau2018}, SLAKE~\cite{Liu2021}, OMVQA-Rad~\cite{Hu2024}, DME-VQA~\cite{TasconMorales2022}, OMVQA-Oph~\cite{Hu2024}, PathMMU~\cite{Sun2024a}, PathVQA~\cite{He2020}, and Quilt-VQA~\cite{Seyfioglu2024}. Medical report generation datasets include MIMIC-CXR~\cite{Johnson2019}, IU-Xray~\cite{DemnerFushman2015}, and Harvard-FairVLMed~\cite{Luo2024}.  We have excluded any retrieval report samples overlapping with these datasets using the image perceptual hashing and carefully checked to ensure no overlap between them. This guarantees that the dataset samples are unseen during ModCLIPs' training and prevents the retrieval results from containing instances identical to the samples. \textbf{Note that the performance on OMVQA-Rad, OMVQA-Oph, and Quilt-VQA can be seen as out-of-distribution results, as they do not include a training split.} Additional dataset details are provided in Appendix~\ref{app:dataset}.

For evaluation metrics, accuracy is used for medical VQA tasks. Radiology report generation is evaluated using BLEU\footnote{In this work, BLEU refers to the average of the BLEU-1, BLEU-2, BLEU-3, and BLEU-4 scores.}~\cite{Papineni2002}, ROUGE-L~\cite{Lin2004}, and RaTEScore~\cite{Zhao2024}, while ophthalmology reports are evaluated using BLEU, ROUGE-L, and METEOR\footnote{RaTEScore is not used for evaluating ophthalmology report generation, as it is specifically for radiology.}~\cite{Banerjee2005}. 

Implementation details regarding report and corpus retrieval, as well as the training of the HeteroRAG framework, are provided in Appendix~\ref{app:implementation}.

\paragraph{Baselines.}

Four categories of baselines are introduced: (1) decoding-based methods aiming for improving factuality including Beam Search~\cite{Sutskever2014}, DoLa~\cite{Chuang2024},  VCD~\cite{Leng2024}, AVISC~\cite{Woo2024}, and M3ID~\cite{Favero2024}; (2) report-retrieval methods including MedDr~\cite{He2024}, FactMM-RAG~\cite{Sun2025}, RULE~\cite{Xia2024}, and MMed-RAG~\cite{Xia2025}; (3) document-retrieval methods including MKGF~\cite{Wu2025} and K-LLaVA~\cite{Hamza2025} and (4) a more recent work that retrieves both reports and documents, MIRA~\cite{Wang2025a}. \textbf{To ensure fair comparison, retrievable reports and documents remain consistent across all baselines. Medical CLIPs for report retrieval also remain consistent across all baselines, with the impact of CLIP training data analyzed separately in Section~\ref{sec:clips}.}
We also introduce widely-used Med-LVLMs: LLaVA-Med-7B~\cite{Li2023}, MedGemma-4B~\cite{Sellergren2025}, HuatuoGPT-V-34B~\cite{Chen2024}, HealthGPT-32B~\cite{Lin2025}, and Lingshu-32B~\cite{Xu2025}. More baseline details are shown in Appendix~\ref{app:baseline}.

\begin{figure}[t]
    \centering
    \includegraphics[width=1\linewidth]{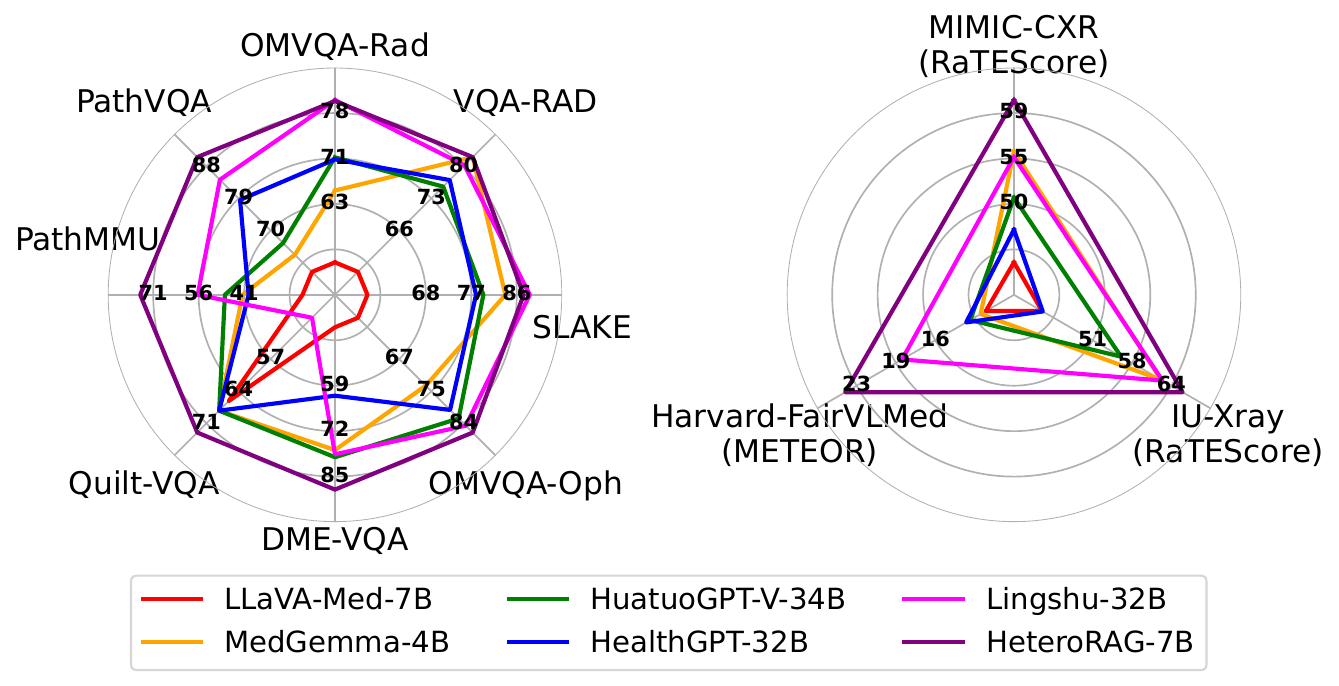}
    \caption{Comparison of HeteroRAG with other Med-LVLMs. Effective retrieval and fine-grained integration of knowledge enables the HeteroRAG to surpass larger Med-LVLMs with greater parameter efficiency.}
    \label{fig:other_lvlm}
\end{figure}

\subsection{Main Results}

The experimental results of different methods based on Lingshu-7B are presented in Table~\ref{tab:main_vqa} and Table~\ref{tab:main_report}. A comparison between widely used Med-LVLMs and HeteroRAG is illustrated in Figure~\ref{fig:other_lvlm}. These results lead to the following key observations: \textbf{(1) Effectiveness of incorporating multi-source knowledge:} HeteroRAG achieves superior performance compared to approaches under different retrieval settings. This demonstrates our effectiveness in retrieving and integrating heterogeneous knowledge. Highly relevant reports enhance the factual accuracy of Med-LVLMs, while evidence documents improve their reliability. \textbf{(2) Generalizability of our framework:} HeteroRAG achieves the best performance on nearly all datasets across three modalities. Notably, this superiority holds not only for closed-ended VQA tasks but also for open-ended report generation and OOD datasets, which require more sophisticated multimodal understanding and generation capabilities. \textbf{(3) Superiority over larger Med-LVLMs:} Figure~\ref{fig:other_lvlm} shows that HeteroRAG, with a 7B parameter size, outperforms most advanced Med-LVLMs, which contain 4 to 5 times more parameters across multiple datasets. This indicates that the proposed framework advances the medical multimodal capabilities of existing Med-LVLMs to a higher level.

\begin{table}[t]
    \centering
    \resizebox{\linewidth}{!}{
    \begin{tabular}{l ccc}
    \toprule[1pt]
    \bf Methods & \bf OMVQA-Rad & \bf OMVQA-Oph & \bf Quilt-VQA \\
    \midrule
Original & 74.92 & 80.83 & 49.27 \\
SFT & 75.00 & 82.17 & 63.85 \\
\textbf{HeteroRAG} & \textbf{80.42} & \textbf{86.00} & \textbf{72.89} \\
\midrule
w/o Reports & 75.08 & 81.33 & 62.97 \\
w/o Doc & 74.17 & 79.08 & 53.94 \\
\midrule
w/o Research & 79.42 & 84.33 & 68.80 \\
w/o Wiki & 77.75 & 84.00 & 67.64 \\
w/o Book & 75.17 & 80.92 & 67.06 \\
w/o Guideline & 79.33 & 84.58 & 69.68 \\
w/o Graph & 78.58 & 83.25 & 66.47 \\
    \bottomrule[1pt]
    \end{tabular}
    }
    \caption{Performance comparison of HeteroRAG against Original and SFT baselines, along with ablation results on various knowledge sources and corpora.}
    \label{tab:ablation_overall}
\end{table}

\begin{table}[t]
    \centering
    \small
    \begin{tabular}{l ccc}
    \toprule[1pt]
    \bf Models & \bf Rad. & \bf Oph. & \bf Pat. \\
    \midrule
CLIP & 6.45 & 2.90 & 5.25 \\
Jina-embeddings-v4 & 8.25 & 4.85 & 6.25 \\
VLM2Vec-V2-2B & 6.00 & 3.05 & 4.45 \\
GME-Qwen2VL-7B & 15.90 & 7.80 & 11.95 \\
Ops-MM-embedding-v1-7B & 16.20 & 9.80 & 15.20 \\
Seed-1.6-embedding & 16.00 & 7.95 & 12.30 \\
    \midrule
BiomedCLIP & 30.20 & 13.45 & 28.85 \\
PMC-CLIP & 30.00 & 19.80 & 23.35 \\
PubMedCLIP* & 13.35 & - & - \\
MM-Retinal* & - & 4.65 & - \\
QuiltNet* & - & - & 39.65 \\
FactMM-RAG* & 44.25 & - & - \\
RULE* & 31.80 & 19.25 & - \\
MMed-RAG* & 31.80 & 19.25 & 30.20 \\
\midrule
\textbf{ModCLIPs*} & \textbf{79.40} & \textbf{47.55} & \textbf{77.35} \\
    \bottomrule[1pt]
    \end{tabular}
    \caption{Image-to-text recall@5 of different retrievers. The asterisks (*) denote the modality-specific retrievers.}
    \label{tab:other_clips}
\end{table}

\begin{table*}[htbp]
    \centering
    \setlength{\tabcolsep}{1mm}
    \resizebox{\linewidth}{!}{
    \begin{tabular}{l c ccc ccc ccc}
    \toprule[1pt]
        \multirow{3}{*}{\textbf{Methods}} & \multirow{3}{*}{\textbf{Retrieval}} & \multicolumn{6}{c}{\textbf{Radiology}} & \multicolumn{3}{c}{\textbf{Ophthalmology}} \\ 
        \cmidrule(r){3-8} \cmidrule(r){9-11}   
        & & \multicolumn{3}{c}{MIMIC-CXR} & \multicolumn{3}{c}{IU-Xray} & \multicolumn{3}{c}{Harvard-FairVLMed} \\
        \cmidrule(r){3-5} \cmidrule(r){6-8} \cmidrule(r){9-11}
        & & BLEU & ROUGE-L & RaTEScore & BLEU & ROUGE-L & RaTEScore & BLEU & ROUGE-L & METEOR \\
        \midrule
Original & - & 10.31 & 30.39 & 53.30 & 18.50 & 41.00 & 57.95 & 4.21 & 14.30 & 15.75 \\
\midrule
Beam Search & - & 10.46 & 30.04 & 50.02 & 24.35 & 42.77 & \underline{65.31} & 2.78 & 11.68 & 13.74 \\
DoLa & - & 10.05 & 30.22 & 52.99 & 18.85 & 40.81 & 58.01 & 5.05 & 16.24 & 18.35 \\
VCD & - & 11.49 & 31.41 & 53.44 & 21.50 & 40.76 & 63.03 & 4.52 & 13.97 & 16.63 \\
AVISC & - & 12.78 & 32.59 & 54.14 & 22.16 & 41.33 & 62.00 & 3.58 & 12.19 & 14.91 \\
M3ID & - & 13.13 & 32.47 & 54.44 & 22.17 & 41.95 & 62.38 & 3.64 & 12.61 & 14.22 \\
\midrule
MedDr & Report & 16.63 & 33.92 & 56.32 & 23.04 & 41.92 & 63.00 & 6.98 & 18.91 & \underline{19.96} \\
FactMM-RAG & Report & 16.94 & 36.03 & 56.86 & 22.74 & 42.93 & 60.70 & 9.42 & 22.58 & 18.40 \\
RULE & Report & 16.49 & 34.02 & 56.47 & 23.95 & 42.69 & 63.51 & 8.39 & 20.94 & 19.91 \\
MMed-RAG & Report & \underline{17.10} & 35.48 & \underline{57.76} & \underline{24.36} & \underline{42.96} & 64.15 & 8.17 & 20.70 & 19.64 \\
\midrule
MKGF & Doc & 11.56 & 32.33 & 53.66 & 19.97 & 41.32 & 59.64 & 5.87 & 16.18 & 16.83 \\
K-LLaVA & Doc & 16.51 & 35.41 & 56.07 & 20.27 & 41.41 & 57.48 & 9.27 & 22.73 & 18.38 \\
\midrule
MIRA & Report+Doc & 16.71 & \underline{36.06} & 57.02 & 19.70 & 41.06 & 57.29 & \underline{9.57} & \underline{22.96} & 18.81 \\
\textbf{HeteroRAG (Ours)} & Report+Doc & \textbf{20.49} & \textbf{39.20} & \textbf{60.73} & \textbf{27.88} & \textbf{46.18} & \textbf{65.92} & \textbf{10.69} & \textbf{24.62} & \textbf{23.63} \\
    \bottomrule[1pt]
    \end{tabular}
    }
    \caption{Model performance of different methods based on Lingshu-7B on the medical report generation task. The best results and second-best results are highlighted in \textbf{bold} and \underline{underlined}, respectively.}
    \label{tab:main_report}
\end{table*}

\begin{table}[t]
    \centering
    \resizebox{\linewidth}{!}{
    \begin{tabular}{l cccc}
    \toprule[1pt]
    \bf Methods & \bf OMVQA-Rad & \bf OMVQA-Oph & \bf Quilt-VQA \\
    \midrule
CLIP & 75.67 & 83.92 & 66.76 \\
\midrule
\textbf{MQG} & \textbf{80.42} & \textbf{86.00} & \textbf{72.89}  \\
w/o DPO & 78.33 & 83.92 & 69.10 \\
w/o SFT & 75.67 & 82.17 & 65.89  \\
    \bottomrule[1pt]
    \end{tabular}
    }
    \caption{Performance of HeteroRAG under two ablation settings: replacing MQG with CLIP-based retrieval, and removing the training stages of MQG.}
    \label{tab:ablation_stage}
\end{table}

\subsection{Effectiveness of Retrieved Knowledge}

We conduct ablation studies to evaluate the contribution of knowledge sources, as shown in Table~\ref{tab:ablation_overall}. The ``Original'' and ``SFT'' settings represent the performance of the original Lingshu-7B and Lingshu-7B after SFT on the original training set, which does not include reports and documents. The other configurations examine HeteroRAG's performance when either reports or documents are removed.
The results show that retrieved knowledge significantly improves Med-LVLM's performance compared to the Original baseline. The performance improvements from supervised fine-tuning alone are insufficient to compensate for the absence of knowledge. When reports or documents are excluded, the performance degradation confirms that both sources are important for HeteroRAG's knowledge-intensive inference. Furthermore, all five corpora contribute to Med-LVLM's capacity.

\subsection{Effectiveness of ModCLIPs}\label{sec:clips}

We evaluate ModCLIPs against other retrievers on image-to-text report retrieval tasks, as shown in Table~\ref{tab:other_clips}. Open-domain retrievers include CLIP~\cite{Radford2021}, Jina-embeddings-v4~\cite{Guenther2025}, VLM2Vec-V2-2B~\cite{Meng2025}, GME-Qwen2VL-7B~\cite{Zhang2024}, Ops-MM-embedding-v1-7B~\cite{ACOAT2025}, and Seed-1.6-embedding~\cite{BST2025}. Generalist medical retrievers include BiomedCLIP and PMC-CLIP~\cite{Lin2023}. Modality-specific medical retrievers include PubMedCLIP~\cite{Eslami2023}, MM-Retinal~\cite{Wu2024}, QuiltNet~\cite{Ikezogwo2023}, FactMM-RAG, RULE and MMed-RAG. For FactMM-RAG, RULE, and MMed-RAG, they are reproduced by fine-tuning BiomedCLIP on their data.

Using the test set described in Section~\ref{sec:method_clips} with recall@5 as our evaluation metric, our experiments demonstrate that ModCLIPs consistently outperform competing methods across all three modalities. This superior performance can be attributed to two key advantages: (1) Single-modality training yields significantly better modality-specific understanding compared to mixed-modality approaches, and (2) our training data offers more comprehensive coverage and greater diversity within each modality.

\subsection{Effectiveness of MQG}

We further investigate the effectiveness of MQG in Table~\ref{tab:ablation_stage}. First, the MQG in HeteroRAG is replaced with a CLIP retrieval module. Specifically, for each medical visual question, the ModCLIPs are employed to retrieve documents through both image-to-text and text-to-text retrieval. The two retrieval results are combined using RRF.
We also ablate the DPO and SFT training stages of the MQG.
The retrieval quality is implicitly evaluated through the downstream performance of Med-LVLMs. We adopt this approach because multimodal medical questions typically lack gold document labels, making classic metrics like MRR or nDCG infeasible. Furthermore, manual assessment of retrieved documents for every query across all baseline methods is impractical due to the large scale of data.
Our experiments demonstrate that MQG retrieves more relevant documents compared to standard CLIP methods. This improvement can be attributed to better alignment of MQG and each corpus's characteristics. Furthermore, both the SFT and DPO training stages prove essential in developing MQG.

\begin{figure}[t]
    \centering
    \includegraphics[width=1\linewidth]{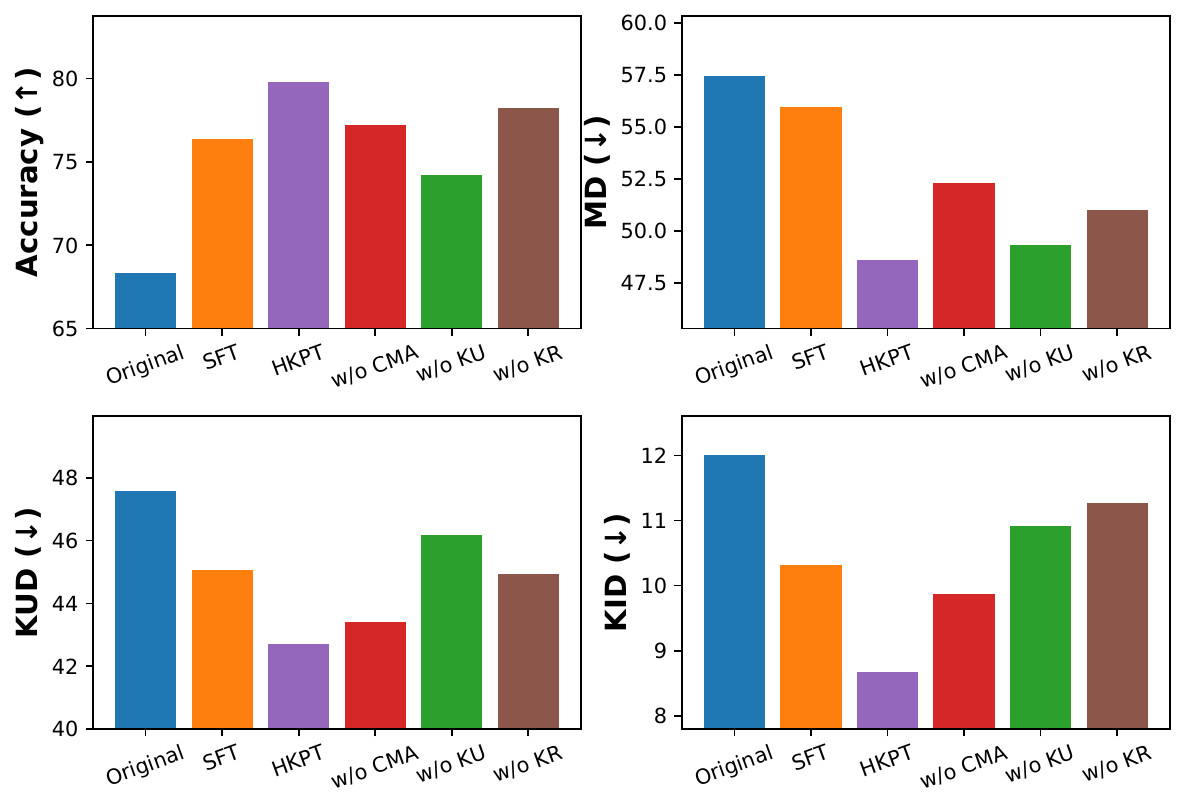}
    \caption{Accuracy and disalignment metrics of Lingshu-7B trained with different methods and data.}
    \label{fig:hkpt_align}
\end{figure}

\subsection{Alignment Effectiveness of HKPT}

To evaluate the alignment effectiveness of HKPT, we introduce three metrics besides answer accuracy: Modality Disalignment (MD), Knowledge Usage Disalignment (KUD), and Knowledge Interference Disalignment (KID). MD corresponds to CMA in Section~\ref{sec:hkpt}, KUD corresponds to KU, and KID corresponds to KR.
MD measures the proportion that the Med-LVLM directly references the retrieved knowledge and correctly answers with the irrelevant image among cases where it originally fails with the irrelevant image.
KUD measures the proportion that the Med-LVLM fails when retrieved contents are introduced among cases where it fails without retrieval.
KID measures the proportion that the Med-LVLM fails when retrieved contents are introduced among cases where it succeeds without retrieval.

Figure~\ref{fig:hkpt_align} shows the average metrics on OMVQA-Rad, OMVQA-Oph, and Quilt-VQA. The ``SFT'' method refers to SFT using the training dataset with documents and reports added. The results demonstrate that HKPT improves overall accuracy compared to both the original and SFT models. Moreover, all three types of disalignment are significantly reduced by the HKPT method. We further conduct ablation studies on each type of preference pair in HKPT, including CMA, KU, and KR. The results confirm that each component contributes effectively to both question answering performance and overall knowledge alignment. Moreover, each component enhances its corresponding alignment capability as expected.

\subsection{Compatibility Analysis}

To analyze the compatibility of the HeteroRAG framework with different Med-LVLMs, we apply it to LLaVA-Med-7B and HuatuoGPT-V-7B besides Lingshu-7B. Specifically, the ModCLIPs and MQG in HRM are kept unchanged, as they are universal across different downstream readers. The HKPT process is performed separately for each Med-LVLM. Results in Table~\ref{tab:compatibility} show that HeteroRAG brings consistent improvements over all Med-LVLMs. This shows the compatibility of the HeteroRAG framework and indicates that HeteroRAG can be transferred to diverse Med-LVLMs.

\begin{table}[t]
    \centering
    \resizebox{\linewidth}{!}{
    \begin{tabular}{l ccc}
    \toprule[1pt]
    \bf Models & \bf OMVQA-Rad & \bf OMVQA-Oph & \bf Quilt-VQA \\
    \midrule
LLaVA-Med-7B & 53.67 & 56.83 & 66.18 \\
+ HeteroRAG & \textbf{59.67} & \textbf{68.42} & \textbf{69.10} \\
\midrule
HuatuoGPT-V-7B & 72.08 & 81.83 & 66.18 \\
+ HeteroRAG & \textbf{76.92} & \textbf{84.33} & \textbf{70.55} \\
\midrule
Lingshu-7B & 74.92 & 80.83 & 49.27 \\
+ HeteroRAG & \textbf{80.42} & \textbf{86.00} & \textbf{72.89} \\
    \bottomrule[1pt]
    \end{tabular}
    }
    \caption{Model performance when the HeteroRAG framework is applied to different Med-LVLMs.}
    \label{tab:compatibility}
\end{table}

\section{Conclusion}

This work addresses the critical challenges of effective retrieval and multi-aspect alignment for heterogeneous knowledge in the Medical MMRAG field. MedAtlas provides a rich, multi-source knowledge base for medical multimodal tasks. The HeteroRAG framework enables precise report retrieval and multi-corpus retrieval, followed by the alignment of heterogeneous knowledge sources. Extensive experiments demonstrate that our framework achieves state-of-the-art performance across multiple medical VQA and report generation benchmarks. Our work paves the way for effectively integrating multi-source knowledge, advancing the reliability of Med-LVLMs in clinical scenarios.

\section*{Limitations}

This work focuses on improving the factual accuracy of Med-LVLMs through effective retrieval and aggregation from heterogeneous medical knowledge sources. Our current framework adopts a one-step retrieval approach. Future work may explore multi-round tool usage and reasoning, such as in OpenAI o3-style systems, to further scale up reasoning capabilities.

Moreover, our primary goal is to enhance factual correctness. Other aspects of practical Med-LVLM deployment, such as fairness, privacy, and safety, are not the focus of this study and remain to be investigated in future work.

\section*{Ethical Consideration}

The multimodal reports in MedAtlas, including radiology, ophthalmology, and pathology subsets, are publicly available and licensed for academic research. For the text corpora in MedAtlas, Research articles, Wiki content, MedQA Textbook, StatPearls, selected clinical guidelines, and Graph data are openly accessible. We will provide access instructions to these publicly available subsets. The remaining materials, including E-books and certain clinical guidelines, can not be directly distributed in their raw form due to data licensing restrictions. However, detailed lists will be publicly available for researchers to obtain these resources.

The medical visual question answering (VQA) and report generation datasets used in this study are publicly available and widely adopted in the research community. The proposed HeteroRAG framework significantly improves the accuracy and reliability of model responses, thereby enhancing the trustworthiness of multimodal medical AI systems. We emphasize responsible data use and follow ethical guidelines for research with publicly available, non-sensitive information.

\section*{Acknowledgements}
This work was supported by the National Natural Science Foundation of China (No. 62576209) and STCSM (No. 2025SHZDZX025G05).

\bibliography{mmrag}

\newpage
\appendix
\section{Additional Analysis}

\begin{figure}[htbp]
    \centering
    \includegraphics[width=0.95\linewidth]{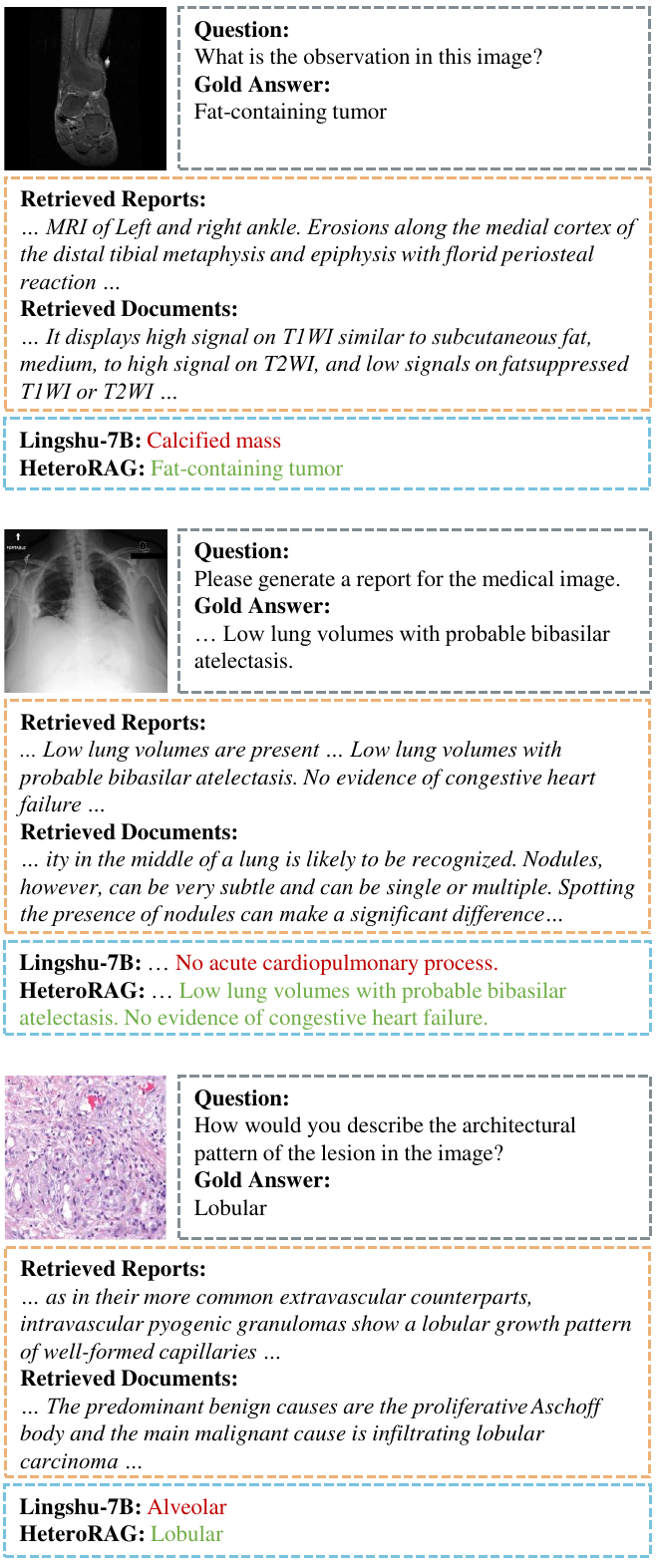}
    \caption{Qualitative analysis I: HeteroRAG effectively leveraging external knowledge.}
    \label{fig:case_study1}
\end{figure}
\begin{figure}[htbp]
    \centering
    \includegraphics[width=0.95\linewidth]{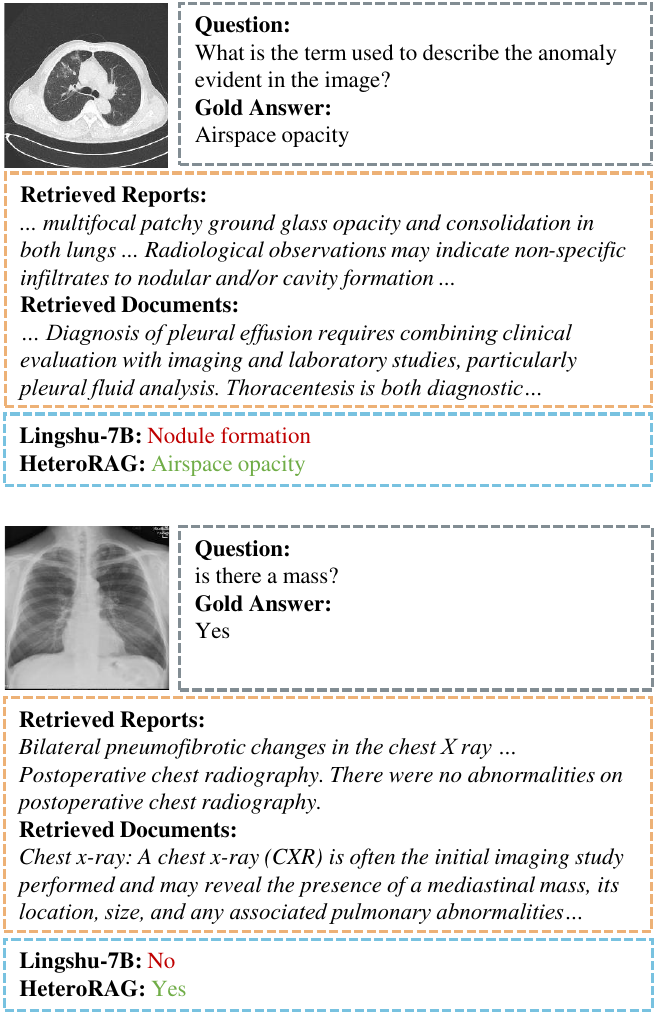}
    \caption{Qualitative analysis II: HeteroRAG resolving conflicting information across multiple knowledge sources.}
    \label{fig:case_study2}
\end{figure}

\subsection{Qualitative Analysis}

\textbf{HeteroRAG significantly improves the factual accuracy of Med-LVLM by effectively leveraging medical reports and documents, as shown in Figure~\ref{fig:case_study1}.} In the first case, HeteroRAG utilizes the retrieved document's description of typical MRI signal characteristics of fat-containing tumors to recognize imaging features indicative of fat content in the lesion, thereby supporting the correct answer.
For the second case, HeteroRAG outperforms Lingshu-7B by effectively leveraging retrieved reports. It refers to key phrases: ``Low lung volumes are present,'' and the impression: ``No evidence of congestive heart failure.'' The similar reports enable clinically accurate conclusions for HeteroRAG.
For the third case, HeteroRAG effectively leveraged both retrieved contents. Retrieved reports state, ``intravascular pyogenic granulomas show a lobular growth pattern of well-formed capillaries,'' where ``lobular growth pattern'' directly corresponds to the ``architectural pattern''. Additionally, the retrieved documents further complement information for ``lobular'', which is a well-established histopathological architectural pattern. HeteroRAG integrated the multi-source knowledge to confirm ``lobular'' as the correct answer.

\textbf{HeteroRAG handles conflicting information acquired from various knowledge sources, as illustrated in Figure~\ref{fig:case_study2}.} In the first case, the image clearly shows airspace opacity, but the retrieved reports contain the misleading term ``nodular formation''. HeteroRAG, via HKPT, successfully integrated other correct evidence (e.g., ``ground glass opacity'') and filtered the conflicting ``nodular'' term, demonstrating robust discrimination in multi-source knowledge competition. The second case features a strong conflict where the image explicitly shows a mass (Answer: Yes), yet some of the retrieved reports state ``no abnormalities''. HeteroRAG successfully rejected this noise, prioritizing the clear visual evidence of the mass.

\subsection{Impact of Retrieved Report Images}

\begin{table}[htbp]
    \centering
    \resizebox{\linewidth}{!}{
    \begin{tabular}{l cccc}
    \toprule[1pt]
    \bf Methods & \bf OMVQA-Rad & \bf OMVQA-Oph & \bf Quilt-VQA & \bf Avg. \\
    \midrule
Original & 74.92 & 80.83 & 49.27 & 68.34 \\
\midrule
HeteroRAG & \textbf{80.42} & \textbf{86.00} & 72.89 & \textbf{79.77} \\
+ Ret. Images & 77.58 & 84.33 & \textbf{73.76} & 78.56 \\
~~~+ Blurring & 78.92 & 85.25 & 72.59 & 78.92 \\
~~~+ Masking & \textbf{80.42} & 85.00 & 72.01 & 79.14 \\
    \bottomrule[1pt]
    \end{tabular}
    }
    \caption{Model performance when the retrieved report images are incorporated.}
    \label{tab:add_report_images}
\end{table}

We further explore the integration of retrieved report images into Med-LVLMs inspired by V-RAG~\cite{Chu2025}. Specifically, we incorporate retrieved report images in constructing the preference pairs and training models. Results in Table~\ref{tab:add_report_images} indicate that adding report images does not improve model performance and even leads to degradation on most datasets.

To further investigate the cause of this degradation, we conduct test-time perturbation experiments. Perturbation settings are introduced: Blurring, which applies a Gaussian blur to the retrieved images to obscure fine-grained visual details; and Masking, which completely obscures the retrieved images with black masks. Results are also shown in Table~\ref{tab:add_report_images}. 

``Blurring'' improves the average performance over the ``+Ret. Images'', and ``Masking'' further restores it to 79.14, close to HeteroRAG without report images (79.77). This validates our hypothesis that visual information in retrieved report images is largely redundant with the report text and hinders the model’s ability to align and integrate external knowledge. Therefore, we exclude retrieved report images in our main methods.

\subsection{HeteroRAG’s Superiority over Larger-Parameter Models}

\begin{table}[t]
    \centering
    \resizebox{\linewidth}{!}{
    \begin{tabular}{l c ccc}
    \toprule[1pt]
    \bf Models & \bf Ret. & \bf OMVQA-Rad & \bf OMVQA-Oph & \bf Quilt-VQA \\
    \midrule
HuatuoGPT-V-34B & N & 71.08 & 82.42 & \textbf{68.22} \\
HealthGPT-32B & N & 70.75 & 80.25 & \textbf{68.22} \\
Lingshu-32B & N & \textbf{80.58} & \textbf{84.25} & 48.40 \\
HeteroRAG-7B & N & 74.33 & 80.58 & 49.85 \\
\midrule
HeteroRAG-7B & Y & 80.42 & 86.00 & 72.89 \\
    \bottomrule[1pt]
    \end{tabular}
    }
    \caption{Model performance when the Retrieval (Ret.) component of HeteroRAG-7B is dropped.}
    \label{tab:openbook}
\end{table}

To investigate why HeteroRAG (built on a 7B backbone) often outperforms 4 to 5 times larger models, we conduct an ablation study by removing its retrieval component. This turns HeteroRAG into a closed-book 7B model. As shown in Table~\ref{tab:openbook}, removing the retrieved knowledge causes a significant performance drop, often resulting in performance weaker than the 4–5× larger models. This proves that the gain is directly attributable to heterogeneous knowledge retrieval and is not merely due to the parametric knowledge of the 7B model. Moreover, HeteroRAG operates as an open-book system that leverages precise external non-parametric knowledge to overcome the inherent capacity limitations of its 7B backbone.

\subsection{Analysis of Query Diversity and Inter-corpus Similarity}

\begin{table*}[t]
\centering
\resizebox{\linewidth}{!}{
\begin{tabular}{lcccccc}
\toprule[1pt]
\textbf{Method} & \textbf{Dist-Research (↑)} & \textbf{Dist-Wiki (↑)} & \textbf{Dist-Book (↑)} & \textbf{Dist-Guideline (↑)} & \textbf{Dist-Graph (↑)} & \textbf{Avg. (↑)}\\
\midrule
Original Question      &  0.1504 & 0.1504 & 0.1504 & 0.1504 & 0.1504 & 0.1504 \\
Zero-shot Rewriting    &  0.2357 & 0.3239 & 0.3048 & 0.3307 & 0.3357 & 0.3062 \\
\textbf{MQG (Ours)}    &  \textbf{0.4105} & \textbf{0.4175} & \textbf{0.3933} & \textbf{0.3936} & \textbf{0.3774} & \textbf{0.3985} 
 \\
\bottomrule[1pt]
\end{tabular}
}
\caption{Query diversity across knowledge sources. Higher values (↑) indicate greater lexical diversity.}
\label{tab:diversity}
\end{table*}

\begin{table*}[t]
\centering
\resizebox{\linewidth}{!}{
\begin{tabular}{lcccccc}
\toprule[1pt]
\textbf{Method} & \textbf{Sim-Research (↓)} & \textbf{Sim-Wiki (↓)} & \textbf{Sim-Book (↓)} & \textbf{Sim-Guideline (↓)} & \textbf{Sim-Graph (↓)} & \textbf{Avg. (↓)}\\
\midrule
Original Question & 1.0000 & 1.0000 & 1.0000 & 1.0000 & 1.0000 & 1.0000 \\
Zero-shot Rewriting & 0.9627 & 0.9656 & 0.9645 & 0.9611 & 0.9156 & 0.9539 \\
\textbf{MQG (Ours)} & \textbf{0.7394} & \textbf{0.7390} & \textbf{0.7358} & \textbf{0.7156} & \textbf{0.7238} & \textbf{0.7307} \\
\bottomrule[1pt]
\end{tabular}
}
\caption{Inter-corpus query similarity across knowledge sources. Lower values (↓) indicate higher query distinctiveness across sources.}
\label{tab:query_sim}
\end{table*}

To assess the diversity of queries from different methods, we calculate the ratio of unique $n$-grams to total generated $n$-grams across the entire test set. We hope to prevent query collapse, where the model generates the same query term for different samples. The metric is formulated as:
\begin{align}
\text{Dist-corpus} &= \frac{1}{2} \sum_{n=1}^{2} \text{Distinct-}n, \label{eq:dist} \\
\text{Distinct-}n &= \frac{\text{Count}(\text{unique } n\text{-grams})}{\text{Count}(\text{total } n\text{-grams})}. \label{eq:distinct-n}
\end{align}
Table \ref{tab:diversity} reports the average metric across OMVQA-Rad, OMVQA-Oph, and Quilt-VQA for different baseline methods. Overall, we can observe that MQG significantly improves the diversity of query terms compared to baseline methods.

Furthermore, we also measure the inter-corpus similarity of queries from different methods. In particular, we measure how similar the queries for different sources are to each other for the same input to avoid query redundancy among queries from different sources. MedCPT-Query-Encoder is used to encode the query for each source and calculate the average cosine similarity of each source's query with the other four sources' queries, respectively. The metric is formulated as:
\begin{equation}
\text{Sim-corpus} = \frac{1}{M-1} \sum_{j=1, j \neq i}^{M} \text{cos}(q_i, q_j),
\end{equation}
where $q_i$ is the encoded query for source $i$, $M$ is the total number of sources (here $M=5$), and $\text{cos}$ denotes the cosine similarity between the two encoded queries.
Table~\ref{tab:query_sim} shows the average metric across OMVQA-Rad, OMVQA-Oph, and Quilt-VQA for different baseline methods. It can be observed that MQG greatly reduces the inter-corpus query similarity, avoiding redundancy among the source queries.

\subsection{Joint Error Analysis of Retrieval and Answer Generation}
\label{subsec:error_analysis}

To analyze cases where the model fails or misuses knowledge sources, and to identify whether the primary bottlenecks reside in the retrieval phase or in the alignment of the Med-LVLM, we conduct a comprehensive joint error analysis. We define four typical error types from the perspectives of input, perception, and integration. First, \textit{Retrieval Failure (Type a)} occurs when the retrieved contents lack the necessary factual evidence to support the gold answer. Second, \textit{Visual Perception Limitation (Type b)} arises when a question requires precise spatial orientations or fine-grained visual features that exceed the model's inherent visual perception capacity, even if the retrieved content is sufficient. Third, \textit{Knowledge Alignment Failure (Type c)} describes instances where both textual evidence and visual cues are explicit, but the model fails to successfully integrate them or is misled. Finally, remaining anomalies are categorized as \textit{Others (Type d)}. The detailed prompt is shown in \ref{prompt:error_analysis}.

Based on these failure types, we annotated error cases from three representative out-of-domain datasets: OMVQA-Rad, OMVQA-Oph, and Quilt-VQA. To ensure reliable and scalable evaluation, we employed the expert medical model in the main method for automated annotation. The quantitative results are summarized in Table \ref{tab:error_analysis}.

\begin{table}[t]
  \centering
    \resizebox{\linewidth}{!}{
  \begin{tabular}{lcccc}
    \toprule
    \textbf{Failure Type} & \textbf{OMVQA-Rad} & \textbf{OMVQA-Oph} & \textbf{Quilt-VQA} & \textbf{Average} \\
    \midrule
    Type a (\%) & 19.57 & 23.21 & \textbf{44.09} & 28.96 \\
    Type b (\%) & 3.40  & 4.76  & 10.75 & 6.30  \\
    Type c (\%) & \textbf{75.74} & \textbf{70.24} & \textbf{44.09} & \textbf{63.36} \\
    \bottomrule
  \end{tabular}
  }
  \caption{Distribution of failure types.}
  \label{tab:error_analysis}
\end{table}

Based on these results, we draw three key conclusions. First, the system bottleneck has successfully shifted from information retrieval to knowledge reasoning; as ModCLIPs and MQG minimize retrieval failures (Type a), alignment and integration (Type c) naturally become the primary challenge. Second, deep medical alignment represents a challenging ``last-mile'' problem for Med-LVLMs, because adjudicating subtle logic conflicts between internal and external heterogeneous knowledge remains difficult for 7B-scale backbones, even with HKPT mitigating disalignment. Finally, heterogeneous knowledge alignment is the core key to medical AI reliability; while high-quality retrieval is a prerequisite, the most critical gains stem from enhancing multi-source alignment through methods like HKPT, pointing the way for future medical AI development.

\subsection{Analysis of Computational Cost}

For the training phase, we focused primarily on the offline data construction stage for MQG training, as the overhead for query generation, document retrieval, and expert model judging constitutes the majority of the HeteroRAG construction time. We also measured the costs for the online inference stage. Specifically, we randomly sampled 200 instances from the training set to measure the time spent on retrieval and inference for data construction. For testing, we sampled 200 VQA and 200 report generation samples from the test sets. Table~\ref{tab:latency_analysis} shows the measurement results.

\begin{table*}[t]
    \centering
    \resizebox{\linewidth}{!}{
    \begin{tabular}{lcc}
        \toprule
        \textbf{Stage} & \textbf{Per-sample Retrieval Latency (s)} & \textbf{Per-sample Inference Latency (s)} \\
        \midrule
        Offline Data Construction          & 4.2 & 17.4 \\
        Online Testing (VQA)               & 0.5 & 1.4  \\
        Online Testing (Report Generation) & 0.4 & 2.8  \\
        \bottomrule
    \end{tabular}
    }
    \caption{Latency analysis across different stages.}
    \label{tab:latency_analysis}
\end{table*}

\section{Additional Details}

\begin{table}[htbp]
    \centering
    \small
    \begin{tabular}{l cccc}
    \toprule[1pt]
        \textbf{Source} & \textbf{Modality} & \textbf{\# Pairs} & \textbf{\# Total} \\
        \midrule
IU-Xray & \multirow{6}{*}{\makecell[c]{Rad.}} & 495 & \multirow{6}{*}{\makecell[c]{1.1M}} \\
PLA & & 14.7k & \\
CheXpert-Plus & & 187.6k & \\
MIMIC-CXR & & 209.6k & \\
ROCOv2 & & 79.8k & \\
PMC-OA-Rad & & 612.2k & \\
        \midrule
Harvard-FairVLMed & \multirow{5}{*}{\makecell[c]{Oph.}} & 5.0k & \multirow{5}{*}{\makecell[c]{112.0k}} \\
DeepEyeNet & & 2.9k & \\
FFA-IR & & 44.7k & \\
MM-Retinal & & 4.4k & \\
PMC-OA-Oph & & 55.1k & \\
        \midrule
ARCH & \multirow{5}{*}{\makecell[c]{Pat.}} & 6.8k & \multirow{5}{*}{\makecell[c]{1.5M}} \\
PathCap & & 221.3k & \\
PatchGastric & & 262.8k & \\
Quilt-1M & & 433.9k & \\
PMC-OA-Pat & & 589.3k & \\
    \bottomrule[1pt]
    \end{tabular}
    \caption{Statistics of multimodal report knowledge base in MedAtlas.}
    \label{tab:stat_report}
\end{table}

\begin{table}[t]
    \centering
    \small
    \setlength{\tabcolsep}{1mm}
    \begin{tabular}{l cccc}
    \toprule[1pt]
        \textbf{Source} & \textbf{Corpus} & \textbf{\# Chunks} & \textbf{\# Total} \\
        \midrule
PubMed & Research & 48.0M & 48.0M \\
        \midrule
Wikipedia & Wiki & 29.7M & 29.7M \\
        \midrule
E-books & \multirow{3}{*}{\makecell[c]{Book}} & 13.7M & \multirow{3}{*}{\makecell[c]{14.1M}} \\
MedQA & & 125.8k & \\
StatPearls &  & 322.7k & \\
        \midrule
Meditron & Guideline & 657.9k & 657.9k \\
        \midrule
        - & - & \textbf{\# Terms} & \textbf{\# Relations} \\
        \midrule
UMLS & Graph & 1.7M & 2.9M \\
    \bottomrule[1pt]
    \end{tabular}
    \caption{Statistics of textual corpora in MedAtlas.}
    \label{tab:stat_corpus}
\end{table}

\subsection{MedAtlas Details}\label{app:medatlas}

The statistics of the multimodal report knowledge base and textual corpora in MedAtlas are shown in Table~\ref{tab:stat_report} and Table~\ref{tab:stat_corpus}, respectively.
For the multimodal report knowledge base, its radiology subset includes IU-Xray~\cite{DemnerFushman2015}, PLA~\cite{Li2024}, CheXpert-Plus~\cite{Chambon2024}, MIMIC-CXR~\cite{Johnson2019}, ROCOv2~\cite{Rueckert2024}, and PMC-OA-Rad~\cite{Lin2023}. The ophthalmology subset includes Harvard-FairVLMed~\cite{Luo2024}, DeepEyeNet~\cite{Huang2021}, FFA-IR~\cite{Li2021}, MM-Retinal~\cite{Wu2024}, and PMC-OA-Oph~\cite{Lin2023}. The pathology subset includes ARCH~\cite{Gamper2021}, PathCap~\cite{Sun2024}, PatchGastric~\cite{Tsuneki2022}, Quilt-1M~\cite{Ikezogwo2023}, and PMC-OA-Pat~\cite{Lin2023}.

The textual knowledge base of MedAtlas encompasses a diverse collection of biomedical and general-domain corpora. The Research corpus includes PubMed Annual Baseline~\cite{NCBI2025}, a comprehensive collection of biomedical literature. The Wiki corpus includes Wikipedia~\cite{Foundation}, providing broad-domain textual knowledge. The Book corpus comprises E-books~\cite{Chen2025, Wu2024a}, MedQA Textbooks~\cite{Jin2020}, and StatPearls~\cite{StatPearls2024}, offering in-depth medical knowledge from authoritative sources. The Guideline corpus includes Meditron Guidelines~\cite{Chen2023}, which contains curated clinical practice guidelines. The Graph corpus is from UMLS Metathesaurus~\cite{Bodenreider2004}, a comprehensive semantic network that integrates concepts and relationships from multiple biomedical vocabularies.

\subsection{Dataset Details}\label{app:dataset}

The datasets used in our work include medical VQA datasets and medical report generation datasets. The VQA datasets are introduced as follows:
\begin{itemize}
    \item \textbf{VQA-RAD}~\cite{Lau2018} is the first manually curated VQA dataset in radiology, where clinical questions were naturally formulated by medical professionals based on radiological images, along with reference answers.
    We employ the closed-ended subset.
    We use the official training split of size 1,027 and the official test split of size 272.
    \item \textbf{SLAKE}~\cite{Liu2021} is a large bilingual medical VQA dataset featuring comprehensive semantic annotations by experienced physicians, accompanied by a structured medical knowledge base.
    We employ the English closed-ended subset.
    We use the official training split of size 1,943 and the official test split of size 416.
    \item \textbf{OMVQA-Rad}~\cite{Hu2024} is the radiology subset of the OmniMedVQA dataset, which aggregates data from multiple medical classification datasets and converts them into a VQA format.
    We employ the open-access subset.
    We randomly select 1,200 samples for the test set.
    \item \textbf{DME-VQA}~\cite{TasconMorales2022} is built upon two public retinal image datasets, IDRiD~\cite{Porwal2018} and e-Ophta~\cite{Decenciere2013}, containing questions related to diabetic macular edema (DME) and other eye conditions. The contours of the original image masks are extracted and rendered as red outlines on the original images to form the question images for each sample.
    We randomly select 5,000 samples from the official training split for the training set and use the official test split of size 1,311.
    \item \textbf{OMVQA-Oph}~\cite{Hu2024} is the ophthalmology subset derived from the OmniMedVQA dataset.
    We employ the open-access subset.
    We randomly select 1,200 samples for the test set.
    \item \textbf{PathMMU}~\cite{Sun2024a} is a high-quality, diverse pathology VQA dataset designed to assess the reasoning and understanding capabilities of large multimodal models in pathology.
    We employ its PathCLS and Atlas subsets, as they are not included in the pretraining data of Lingshu-7B to the best of our knowledge.
    Then we randomly select 2,095 samples for the training set and 598 samples for the test set.
    \item \textbf{PathVQA}~\cite{He2020} is the first VQA dataset in pathology, constructed using a semi-automated pipeline that extracts question-answer pairs from pathology textbooks and digital libraries.
    We employ the closed-ended subset.
    We randomly select 5,000 samples from the official training split for the training set and use the official test split of size 3,391.
    \item \textbf{Quilt-VQA}~\cite{Seyfioglu2024} is an organic evaluation dataset created by extracting real-world medical questions and answers from QUILT educational videos.
    We employ the closed-ended subset.
    We randomly select 343 samples for the test set.
\end{itemize}

The medical report generation datasets are described as follows:
\begin{itemize}
    \item \textbf{MIMIC-CXR}~\cite{Johnson2019} is a large, publicly available collection of chest radiographs in DICOM format, paired with free-text radiology reports from studies conducted at the Beth Israel Deaconess Medical Center in Boston, MA.
    We exclude the samples that do not contain findings or impressions.
    We randomly select 5,000 samples from the official training split for the training set and use the official test split of size 1,624.
    \item \textbf{IU-Xray}~\cite{DemnerFushman2015} consists of chest X-ray images linked to their corresponding clinical diagnostic reports.
    We exclude the samples that do not contain findings or impressions.
    We use the official training split of size 2,445 and the official test split of size 296.
    \item \textbf{Harvard-FairVLMed}~\cite{Luo2024} includes patient records with SLO fundus images and clinical notes for glaucoma diagnosis.
    We randomly select 3,500 samples from the official training split for the training set and 1,000 samples from the official test split for the test set.
\end{itemize}

\subsection{Baseline Details}\label{app:baseline}

Decoding-based methods aiming to improve factuality are described as follows:
\begin{itemize}
    \item \textbf{Original} uses greedy decoding, which selects the token with the highest probability at each generation step, favoring locally optimal choices without considering long-term sequence quality.
    \item \textbf{Beam Search}~\cite{Sutskever2014} improves upon greedy decoding by keeping track of multiple partial sequences (beams) at each step, exploring a wider range of potential outputs and often yielding more coherent and accurate generations.
    \item \textbf{DoLa}~\cite{Chuang2024} leverages the discrepancy between early and later layer representations in the model by comparing their projected logits onto the vocabulary space, guiding generation toward more accurate and contextually appropriate tokens.
    \item \textbf{VCD}~\cite{Leng2024} introduces a training-free decoding strategy that compares outputs from original and perturbed visual inputs, helping to mitigate model reliance on statistical bias and unimodal priors.
    \item \textbf{AVISC}~\cite{Woo2024} is a test-time decoding method that enhances visual understanding by dynamically recalibrating attention during token generation, specifically reducing over-attention to image tokens that lack task-relevant content.
    \item \textbf{M3ID}~\cite{Favero2024} strengthens the impact of the reference image during generation by amplifying tokens that have higher mutual information with the visual input.
\end{itemize}

Medical report-retrieval methods are described as follows:
\begin{itemize}
    \item \textbf{MedDr}~\cite{He2024} employs a retrieval-augmented medical diagnosis strategy in the inference process to improve the factuality of the model's responses.
    \item \textbf{FactMM-RAG}~\cite{Sun2025} feeds the multimodal question together with the retrieved report to the Med-LVLM, which is fine-tuned using standard SFT to better incorporate external reports.
    \item \textbf{RULE}~\cite{Xia2024} constructs a preference dataset focusing on cases where over-reliance on retrieved reports causes errors, aiming to balance the use of internal knowledge and external context.
    \item \textbf{MMed-RAG}~\cite{Xia2025} extends RULE~\cite{Xia2024} by introducing cross-modality alignment to ensure image utilization and proposing overall alignment to better incorporate external reports.
\end{itemize}

Medical document-retrieval methods are described as follows:
\begin{itemize}
    \item \textbf{MKGF}~\cite{Wu2025} uses a multimodal retriever to fetch knowledge graphs and supplement knowledge for LVLMs. We reproduce it using ModCLIP for image-to-text and text-to-text retrieval to retrieve text corpora, combining results via Reciprocal Rank Fusion.
    \item \textbf{K-LLaVA}~\cite{Hamza2025} retrieves relevant KG triplets using a CLIP model and fine-tunes the LVLM to incorporate the knowledge. We also use ModCLIP for retrieval in this method.
\end{itemize}

A more recent work that retrieves both reports and documents is described as follows:
\begin{itemize}
    \item \textbf{MIRA}~\cite{Wang2025a} is a method that retrieves both medical reports and documents. To reproduce it, we use the input image to retrieve similar clinical cases and employ a zero-shot query-rewriting module~(Lingshu-7B) for corpus retrieval. Then the downstream reader is fine-tuned, whose training data includes a chain-of-thought to guide the reader in analyzing the external knowledge.
\end{itemize}

We also introduce widely used Med-LVLMs, which are described as follows:
\begin{itemize}
    \item \textbf{LLaVA-Med-7B}~\cite{Li2023} first aligns biomedical terminology using figure-caption pairs from scientific literature, then enhances conversational understanding through GPT-4-generated instruction-following data, simulating the way non-experts gradually acquire medical knowledge through.
    \item \textbf{MedGemma-4B}~\cite{Sellergren2025} is developed by Google and exhibits strong medical image and text understanding capabilities, significantly outperforming other generative models of similar size and approaching the performance of specialized task-specific models.
    \item \textbf{HuatuoGPT-V-34B}~\cite{Chen2024} is trained on PubMedVision, a large-scale dataset of 1.3 million medical VQA samples constructed by refining image-text pairs from PubMed with the help of MLLMs (e.g., GPT-4V), showing superior performance in medical multimodal scenarios.
    \item \textbf{HealthGPT-32B}~\cite{Lin2025} integrates medical visual comprehension and generation into a unified autoregressive framework, progressively adapting heterogeneous multimodal knowledge to a pre-trained LLM through a bootstrapping approach.
    \item \textbf{Lingshu-32B}~\cite{Xu2025} is developed based on a carefully curated multimodal dataset enriched with comprehensive medical knowledge, undergoing multi-stage training to progressively embed domain expertise and improve task-solving abilities, consistently outperforming existing open-source models in most medical multimodal benchmarks.
\end{itemize}

\subsection{Implementation Details}\label{app:implementation}

For report retrieval, we adopt the adaptive retrieval context selection method \cite{Xia2025}. For document retrieval, the MQG generates up to four queries in total for unstructured corpora (all except Graph). Each query retrieves the top-10 documents, which are then reranked to select the top-2 documents. For the Graph corpus, the MQG retrieves one term and re-ranks the top-10 relations. The correctness threshold $\alpha_r$ in HKPT is set to 50/3 for radiology reports and 20/3 for ophthalmology reports, which is tuned based on the external development set.

For the training of ModCLIPs, they are initialized from BiomedCLIP~\cite{Zhang2023}. The learning rate is set to 2e-4, and the batch size is set to 512. The number of training epochs of radiology, ophthalmology, and pathology ModCLIP is set to 10, 100, and 10, respectively, for the different sizes of modality image-text pairs.

For the training of MQG, the Med-LVLM is initialized from Lingshu-7B~\cite{Xu2025}. We use LoRA~\cite{Hu2022} for efficient fine-tuning. For the SFT process, its learning rate is set to 2e-4, the batch size is set to 64, and the number of epochs is 3. For the DPO process, its learning rate is set to 2e-5, the batch size is set to 64, and the number of epochs is set to 3. 

For the training of HKPT, the Med-LVLM is initialized from Lingshu-7B. We also use LoRA~\cite{Hu2022} for efficient fine-tuning. Its learning rate is set to 2e-5, the batch size is set to 64, and the number of epochs is set to 3.

In our experiments, we use the development set, which has no overlap with the training and test sets, to tune the hyperparameters. We use the AdamW~\cite{Loshchilov2019} as our optimizer. For the test stage, the temperature is set to 0 to ensure reproducibility. Detailed prompts are provided in Appendix~\ref{app:prompt}. Huggingface Trainer is adopted as the training framework for Med-LVLMs.

\section{AI Assistance Statement}

LLM tools are employed solely for language refinement and grammar correction to improve overall clarity and readability.

\section{Prompt List}\label{app:prompt}

\begin{prompt}[title={Prompt \thetcbcounter: VQA with Retrieved Reports and Documents}, label=prompt:vqa, breakable]
\{question\_image\}\\
\\
Retrieved Contents:\\
\{text\_doc\}\\
\\
Reference Reports:\\
\{mm\_doc\}\\
\\
\{question\_text\}\\
Please answer the question based on the Retrieved Contents. It should be noted that the diagnostic information in the Reference Reports cannot be directly used as the basis for diagnosis, but should only be used for reference and comparison.\\
\\
Answer with the option's letter from the given choices directly.
\end{prompt}

\begin{prompt}[title={Prompt \thetcbcounter: Report Generation with Retrieved Reports and Documents}, label=prompt:report_generation, breakable]
\{question\_image\}\\
\\
Retrieved Contents:\\
\{text\_doc\}\\
\\
Reference Reports:\\
\{mm\_doc\}\\
\\

Please answer the question based on the Retrieved Contents. It should be noted that the diagnostic information in the Reference Reports cannot be directly used as the basis for diagnosis, but should only be used for reference and comparison.\\
\\
(For radiology) You are a helpful assistant. Please generate a report for the given image, including both findings and impressions. Return the report in the following format: Findings: \{\} Impression: \{\}.\\
(For ophthalmology) You are a helpful assistant. Please generate a short report for the given image in 100 words. Please only include the content of the report in your response.
\end{prompt}

\begin{prompt}[title={Prompt \thetcbcounter: Query Exploration by the Expert Med-LVLM}, label=prompt:query_exploration, breakable]
\{question\_image\}\\
\\
\# Question (based on the image)\\
\{question\_text\}\\
\\
\# Corpus Description\\
research: The corpus provides access to advanced biomedical research, facilitating access to specialized knowledge and resources.\\
wiki: The corpus provides access to general knowledge across a wide range of topics.\\
book: The corpus provides access to medical knowledge resource including various educational resources and textbooks.\\
guideline: The corpus provides access to clinical guidelines from leading health organizations.\\
graph: The corpus provides a structured knowledge graph that connects medical definitions and related terms.\\
\\
\# Query Format\\
$<$research$>$\{query0\} ; \{query1\} ; ... (Use ; to separate the queries)$<$/research$>$\\
$<$wiki$>$\{query0\} ; \{query1\} ; ... (Use ; to separate the queries)$<$/wiki$>$\\
$<$book$>$\{query0\} ; \{query1\} ; ... (Use ; to separate the queries)$<$/book$>$\\
$<$guideline$>$\{query0\} ; \{query1\} ; ... (Use ; to separate the queries)$<$/guideline$>$\\
$<$graph$>$\{query\_term0\} , \{query\_relation0\} ; \{query\_term1\} , \{query\_relation1\} ; ... (Use ; to separate the queries. Each query should use , to separate the \{query\_term\} and \{query\_relation\})$<$/graph$>$\\
\\
To answer the question labeled as \# Question, please construct appropriate queries to get the information you need.\\
1. Each corpus in \# Corpus Description must have search queries constructed.\\
2. Please give the search queries following the format in \# Query Format. Each corpus should have 6 queries, separated by ';'.\\
3. The queries generated for each corpus should exhibit diversity and be closely aligned with the specific information needs and characteristics of that corpus.
\end{prompt}

\begin{prompt}[title={Prompt \thetcbcounter: Query Judging through Retrieved Documents by the Expert Med-LVLM}, label=prompt:query_judging, breakable]
\{question\_image\}\\
\\
\# Question (based on the image)\\
\{question\_text\}\\
\\
\# Gold Answer\\
\{gold\}\\
\\
\# Documents\\
\{documents\}\\
\\
You are a professional medical expert. Please judge whether the information in the \# Documents supports the \# Gold Answer as a response to the \# Question. Please judge whether \# Documents supports the \# Gold Answer in response to the \# Question, rather than evaluating if the \# Question's answer is the \# Gold Answer. Please first think step-by-step and then show your judgement using the format $<$answer$>$yes/no$<$/answer$>$ at the end of your response. Please keep your entire response simple and complete, up to 100 words.
\end{prompt}

\begin{prompt}[title={Prompt \thetcbcounter: Query Generation by the Multi-corpora Query Generator}, label=prompt:query_generation, breakable]
\{question\_image\}\\
\\
\# Question (based on the image)\\
\{question\_text\}\\
\\
\# Corpus Description\\
research: The corpus provides access to advanced biomedical research, facilitating access to specialized knowledge and resources.\\
wiki: The corpus provides access to general knowledge across a wide range of topics.\\
book: The corpus provides access to medical knowledge resource including various educational resources and textbooks.\\
guideline: The corpus provides access to clinical guidelines from leading health organizations.\\
graph: The corpus provides a structured knowledge graph that connects medical definitions and related terms.\\
\\
\# Query Format\\
$<$research$>$\{query\}$<$/research$>$\\
$<$wiki$>$\{query\}$<$/wiki$>$\\
$<$book$>$\{query\}$<$/book$>$\\
$<$guideline$>$\{query\}$<$/guideline$>$\\
$<$graph$>$\{query\_term\} , \{query\_relation\} (Each query should use , to separate the \{query\_term\} and \{query\_relation\})$<$/graph$>$\\
\\
To answer the question labeled as \# Question, please construct appropriate queries to get the information you need.\\
1. Please give the search queries following the format in \# Query Format. For each corpus, if you think no information retrieval is needed, simply output an empty tag for that corpus, for example: $<$book$>$$<$/book$>$.\\
2. The queries generated for each corpus should be closely aligned with the specific information needs and characteristics of that corpus.
\end{prompt}

\begin{prompt}[title={Prompt \thetcbcounter: Joint Error Analysis of Retrieval and Answer Generation}, label=prompt:error_analysis, breakable]
\{question\_image\}\\
\\
\# Retrieved Contents\\
\{reports\_and\_documents\}\\
\\
\# Question (based on the image)\\
\{question\_text\}\\
\\
\# Gold Answer\\
\{gold\}\\
\\
\# Wrong Response\\
\{wrong\}\\
\\
You are a professional medical expert. The current model has provided an incorrect answer \# Wrong Response instead of the \# Gold Answer to the question \# Question (based on the image) despite having access to \# Retrieved Contents. Categorize the model's failure into exactly one of these four mutually exclusive categories:\\
a. Retrieval Failure: The \# Retrieved Contents lack the necessary factual evidence to support the \# Gold Answer.\\
b. Visual Perception Limitation: Even if the retrieved content is sufficient, the question requires precise spatial orientations or fine-grained visual features that may exceed the model's inherent visual perception capacity.\\
c. Knowledge Alignment Failure: Both textual evidence and visual cues are explicit, but the model failed to integrate them or was misled.\\
d. Others\\
\\
Please first think step-by-step and then show your judgement using the format $<$answer$>$a/b/c/d$<$/answer$>$ at the end of your response. Only one option can be chosen. Please keep your entire response simple and complete, up to 100 words.
\end{prompt}

\end{document}